\documentclass[10pt,twocolumn,letterpaper]{article}

\usepackage{cvpr}
\usepackage{times}
\usepackage{epsfig}
\usepackage{graphicx}
\usepackage{amsmath}
\usepackage{amssymb}
\usepackage{amsfonts}
\usepackage{balance}  
\usepackage{array}
\usepackage{multirow}
\usepackage{makecell}
\usepackage{xcolor}
\usepackage[pagebackref=true,breaklinks=true,letterpaper=true,colorlinks,bookmarks=false]{hyperref}

\usepackage{booktabs,subcaption,amsfonts,dcolumn}
\newcolumntype{d}[1]{D..{#1}}
\usepackage[breaklinks=true,bookmarks=false]{hyperref}

\cvprfinalcopy % *** Uncomment this line for the final submission

\def\cvprPaperID{****} % *** Enter the CVPR Paper ID here

\begin{document}

%%%%%%%%% TITLE
\title{M$^{3}$D-GAN: Multi-Modal Multi-Domain Translation with Universal Attention}

\author{Shuang Ma\\
SUNY Buffalo\\
Buffalo, NY\\
{\tt\small shuangma@buffalo.edu}
% For a paper whose authors are all at the same institution,
% omit the following lines up until the closing ``}''.
% Additional authors and addresses can be added with ``\and'',
% just like the second author.
% To save space, use either the email address or home page, not both
\and
Daniel McDuff\\
Microsoft Research\\
Redmond, WA\\
{\tt\small damcduff@microsoft.com}
\and
Yale Song\\
Microsoft Cognition\\
Redmond, WA\\
{\tt\small yalesong@microsoft.com}
}

\maketitle
%\thispagestyle{empty}

%%%%%%%%% ABSTRACT
\begin{abstract}
Generative adversarial networks have led to significant advances in cross-modal/domain translation. However, typically these networks are designed for a specific task (e.g., dialogue generation or image synthesis, but not both). We present a unified model, M$^{3}$D-GAN, that can translate across a wide range of modalities (e.g., text, image, and speech) and domains (e.g., attributes in images or emotions in speech). Our model consists of modality subnets that convert data from different modalities into unified representations, and a unified computing body where data from different modalities share the same network architecture. We introduce a universal attention module that is jointly trained with the whole network and learns to encode a large range of domain information into a highly structured latent space. We use this to control synthesis in novel ways, such as producing diverse realistic pictures from a sketch or varying the emotion of synthesized speech. We evaluate our approach on extensive benchmark tasks, including image-to-image, text-to-image, image captioning, text-to-speech, speech recognition, and machine translation. Our results show state-of-the-art performance on some of the tasks.
\end{abstract}

%%%%%%%%% BODY TEXT
\begin{figure}[h]
\begin{center}
    \includegraphics[width=1\linewidth]{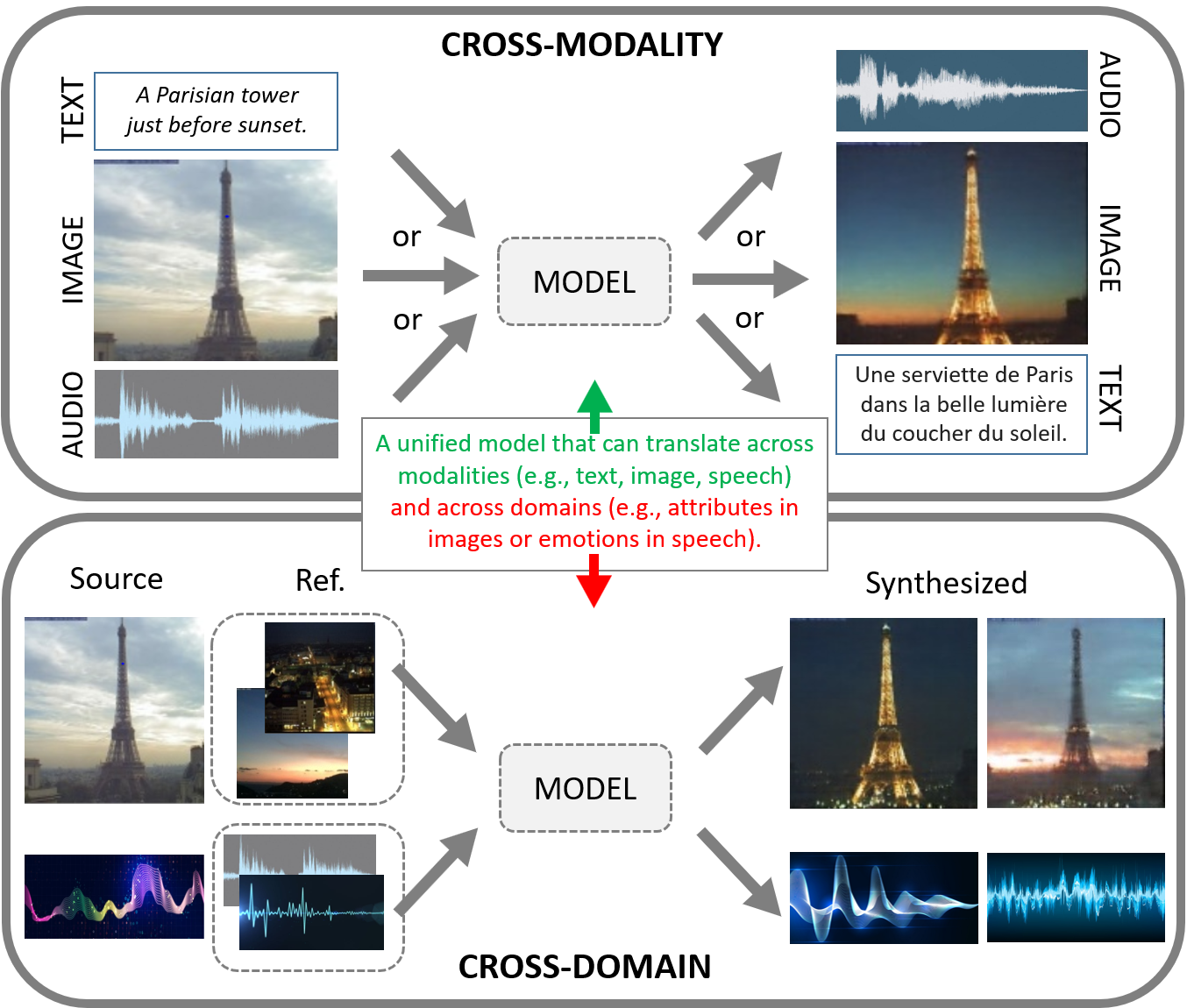}
\end{center}
\vspace{-5pt}
   \caption{We present a unified model that can translate across multiple modalities (synthesize text, images or audio from text, images or audio) and multiple domains (synthesize a diverse set of examples from a single source input with different attributes) in a controllable fashion.}
\label{fig:summary}
\end{figure}

\section{Introduction}
Generative adversarial networks~\cite{GANs} learn a mapping from source to target distributions, and have shown great performances in data translation tasks involving different modalities such as image, text, audio. These problems are typically multi-modal (e.g., text-to-speech or text-to-image), and the mappings are inherently one-to-many (e.g., the same sentence said with different emotions should sound different). Thus, it would be ideal if a model is able to learn a mapping across multiple \textit{modalities} that also allows for the \textit{domain} to be explicitly controlled.

We define each \textit{modality} as a set of data of the same type, and each \textit{domain} as a set of data with the same attribute value. For example, in Fig~\ref{fig:summary}, text would be one modality, images another and audio yet another. Images captured at night represent one domain and images captured in the daytime represent another domain. A unified architecture would make the modeling more efficient and allow representations to be shared, even if they come from different domains or modalities.

In this work, we introduce a multi-modal multi-domain generative adversarial network (M$^{3}$D-GAN) -- a unified model that can translate across a wide range of modalities (e.g. image, text and speech) and domains (e.g., styles and attributes). We specifically focus on the ability to explicitly control the domain aspect, rather than randomly generating results as in the previous work~\cite{Bicycle-GAN}.
Creating such an architecture is non-trivial for several reasons: 
(1) Different models are designed for different data types; it is non-trivial to extend them across different modalities (e.g., a network designed for text-to-image translation would not work for speech-to-text translation).
(2) The mode collapse issue widely observed in GANs makes it difficult to produce diverse results. 
(3) The alternating training process in GANs makes it difficult to explicitly control the domain aspect in the synthesized output. 

Our M$^{3}$D-GAN consists of modality-specific sub-networks and a unified computing body. The former converts input samples from different modalities into unified representations. The majority of computation is done in the unified computing body, where the representations from all modalities share the same network (but not the weights). We achieve the \textit{cross-modality} property by letting different tasks from the same modality share their modality subnets, which avoids creating networks for every tasks. For example, all image translation tasks share the same image-modality-net, no matter which image domains they come from (e.g., shoes or scenes). This encourages generalization across tasks and makes it easy to add additional tasks. 

To produce diverse results, we introduce a \textit{universal attention module} that encodes multi-domain distributions in a highly structured latent space by means of information bottleneck~\cite{tishby2000information}. At inference time, we synthesize diverse samples by providing a condition, or randomly sampling from the latent space induced by this module.

In summary, our contributions are: (1) We propose multi-modal multi-domain GAN (M$^3$D-GAN) that can translate data across multiple modalities and domains. (2) We introduce a universal attention module that encodes multi-domain variations in a latent space. (3) We evaluate our approach on a broad range of tasks and achieve superior results compared to baseline models.

\begin{figure}[t]
\begin{center}
    \includegraphics[width=1\linewidth]{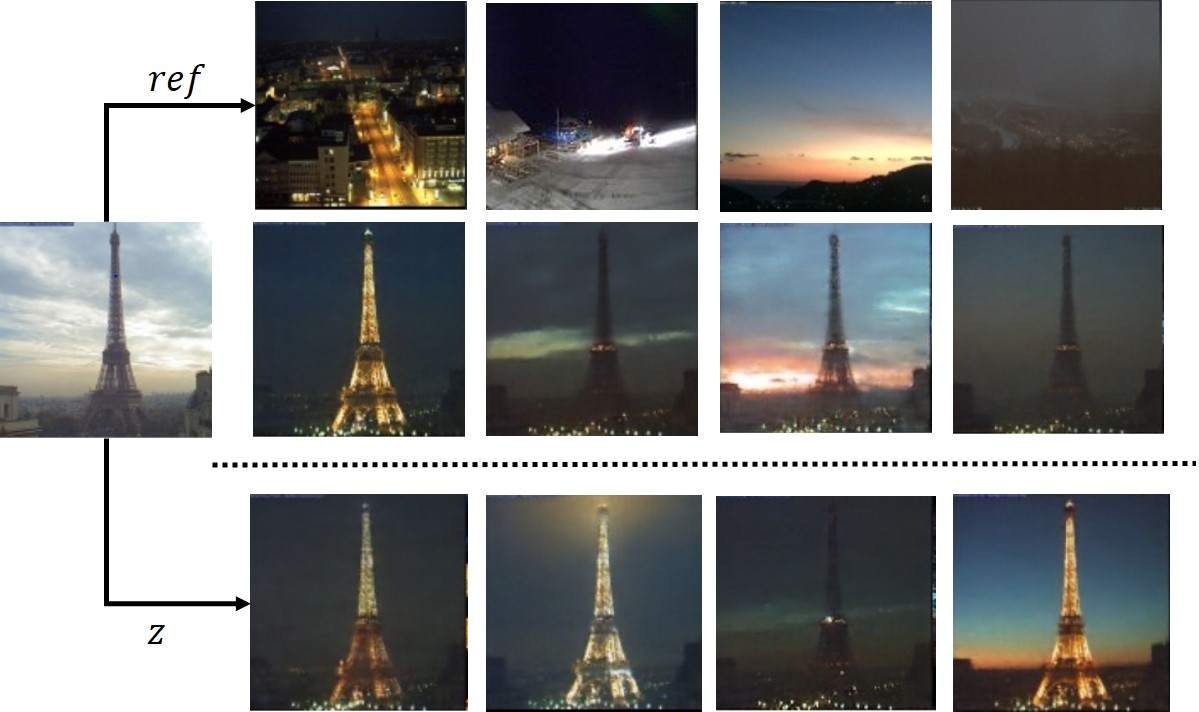}
\end{center}
\vspace{-5pt}
   \caption{Results of our model on image-to-image translation given an input image (left). On the right we show reference images (top), reference-conditioned synthesis results (middle), and unconditional synthesis results. 
   }
\label{fig:paris}
\end{figure}

\begin{figure*}[t]
\begin{center}
    \includegraphics[width=1.0\linewidth]{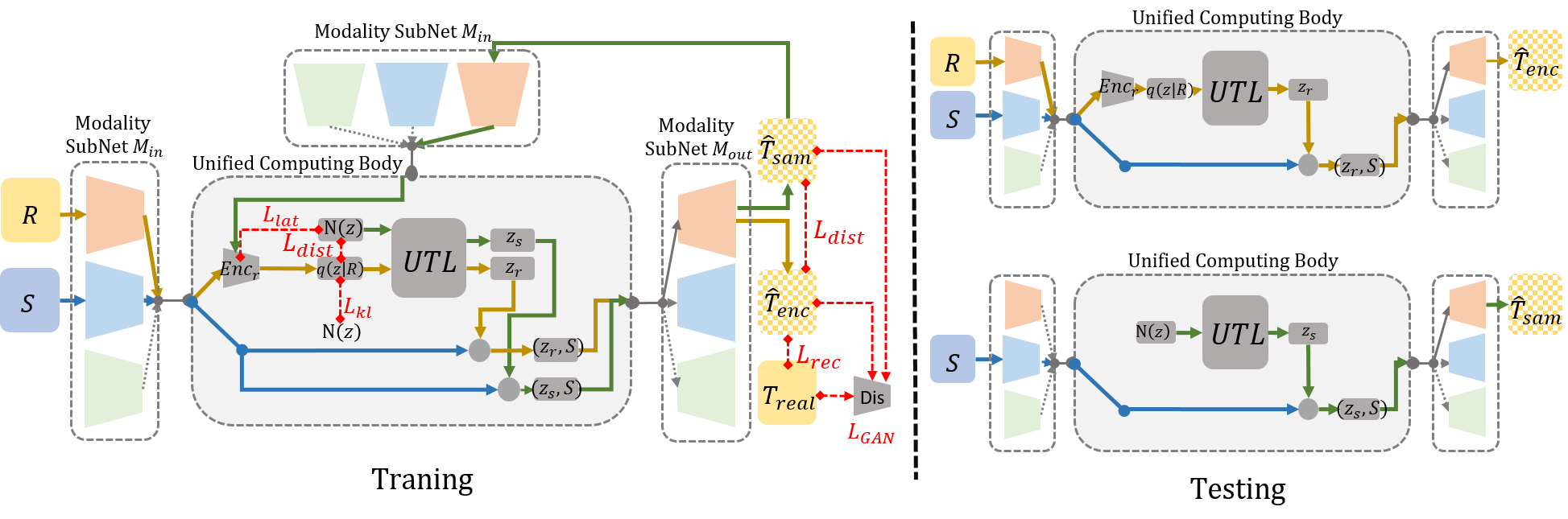}
\end{center}
   \caption{M$^{3}$D-GAN architecture. 
   \textbf{Training:} We use the modality subnets $\mathbf{M}_{in}$ to convert data into a universal representation. These are processed via a universal computing body to produce latent codes $z_s: z \sim \mathcal{N}(0, I)$ and $z_r: z \sim E_r(R)$. We combine these with the source $S$ and feed to the Modality-specific generator ($\mathbf{M}_{out}$) to convert them into the desired modality for synthesis. 
   \textbf{Inference:} Given a source sample $S$, we synthesize a new sample by either \textbf{(Top)} providing a reference $R$ to explicitly control the desired domain, or \textbf{(Bottom)} sampling a latent code from a normal distribution.}
\label{fig:pipeline}
\end{figure*}

\section{Related Work}
\textbf{Cross-Domain Synthesis:} In the image modality, conditioning GANs with labels has been used to generate higher resolution images~\cite{odena2017conditional}. An important aspect of synthesis is the ability to frame the task as the description of attributes or features~\cite{attribute}. Modeling images in layers (i.e, foreground and background) allows for explicit control over the image attributes~\cite{attribute}. Pix2pix~\cite{pix2pix} made the first attempt at translating across different image domains by training on paired data (e.g., sketches to photos). Since then, more image translation work has used unsupervised training~\cite{cycleGAN,cross-domain-image-generation, DA-GAN, distance-GAN}. However, most GAN architectures require retraining for each specific tasks. Bridging the gaps between domains, Choi et al., proposed StarGAN~\cite{choi2017stargan}, which is a unified model for image-to-image translation that produced compelling results for controllable image generation, enabling cross-domain image synthesis.

\textbf{Cross-Modality Synthesis}: GANs have had a significant impact on data synthesis cross modalities, from image-to-text to speech synthesis. Text-to-image generation is an example of cross-domain synthesis~\cite{stackgan,text2image_reed, DA-GAN}. Image captioning can be considered as an inverse process of generating text from images~\cite{show&tell}. Analogously, speech recognition aims to turn a segment of audio into a textual string and similarly deep networks have been successfully employed for this~\cite{zhang2017very,xiong2018microsoft}. Text-to-speech synthesis presents the challenge of modeling the style in which the speech is synthesized~\cite{ma-iclr19}. To this end, the recently proposed Tacotron-based approaches \cite{Tacotron,DBLP:conf/icml/Wang-GST_Tacotron} use a piece of reference speech audio to specify the expected style. This approach means a single textual string can be synthesized in numerous ways (a one to many mapping).  Controlling the attributes of images~\cite{choi2017stargan,attribute} is somewhat similar to this, ideally we would want to control the domain of the synthesized data.  

\section{Approach}
Neuroscience has revealed that the brain forms unified abstract representations from cross-sensory modalities~\cite{pietrini2004beyond,giard1999auditory}. For instance, the same mirror neurons are shown to fire when primates observe actions and hear sounds associated with them~\cite{kohler2002hearing,giard1999auditory}; visual and tactile recognition uses similar processes~\cite{pietrini2004beyond}. 
While there is much we do not understand about the nature of how the brain processes multi-sensory signals, we were inspired by this property in the design of our network. Considering the differences between human senses and perception, our approach consists of a set of modality sub-networks, designed to convert input data into unified representations somewhat analogous to the senses, and a unified computing body to transfer these representations, inspired by the abstracted notion of perception.

Given two data distributions, source $\mathbf{S}$ and target $\mathbf{T}$, we aim to learn a mapping that is robust to different data types (multi-modal) and to find diverse and plausible $t \in \mathbf{T}$ that correspond to $s \in \mathbf{S}$ (multi-domain). We denote the modality subnets as $ \mathbf{M}^j = \{\mathbf{M}_{in}, \mathbf{M}_{out}\}^j$, $j$ = \{text, image, speech\}. $\mathbf{M}_{in}^j$ are prenets, where samples from different modalities are converted to unified representations, $\mathbf{M}_{in}(\mathbf{S}) \to S$ and $\mathbf{M}_{in}(\mathbf{T}) \to T$. The unified computing body learns to relate and transfer the representations from the $S$ to $T$. $\mathbf{M}_{out}^j$ is a modality-specific generator, where the outputs from the unified computing body are fed into it and are translated to the desired data modality.

During training, we have a collection of paired samples ${(s \in \mathbf{S}, t \in \mathbf{T})}$. However, the training dataset usually contains only one such pair. Therefore, we introduce a universal attention module $E_{att}$ to encode diverse domain information from the target distribution into a latent space. At test time, M$^{3}$D-GAN can produce diverse results by either explicitly providing a reference sample as a conditioning variable or randomly sampling from the latent space. For a simple notation, in this paper, we denote the domain-specific generator $\mathbf{M}_{out}^j$ as $G$.
\begin{figure}[t]
\begin{center}
    \includegraphics[width=1\linewidth]{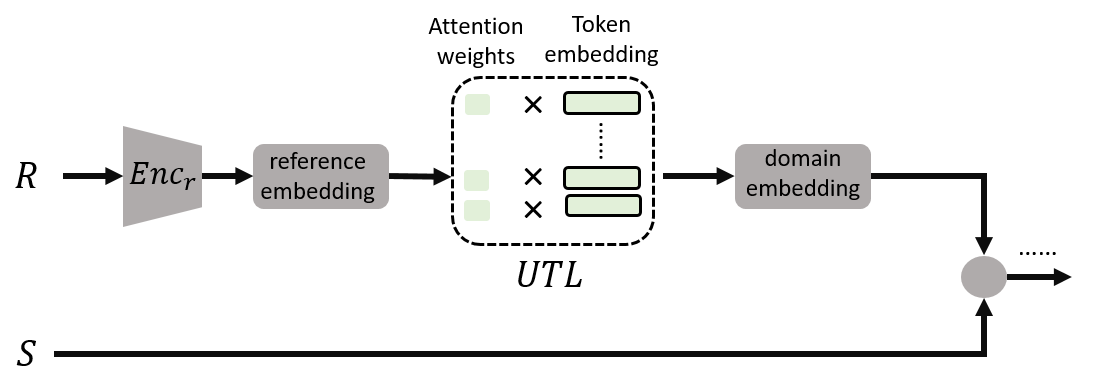}
\end{center}
   \caption{An illustration of our universal attention module.}
\label{fig:atten_module}
\end{figure}

\subsection{Universal Attention Module} \label{sec:attention}
Figure~\ref{fig:atten_module} shows an illustration of our universal attention module. We aim to model a variety of domain information from the target distribution; during training, rather than starting only from a source sample (e.g. pix2pix~\cite{pix2pix}), we also take samples from the target distribution as references, denoted by $R$, i.e. $\{S,R\} \to \hat{T}$. When training on paired data, e.g. \{text, image\}, $R$ is the ground-truth target. For testing, $R$ can be any sample from $\mathbf{T}$ and specifies the domain we wish to synthesize. To this end, we design $E_{att}$ that consists of a reference encoder ($Enc_r$) and a universal token layer ($UTL$). They are jointly trained with the whole model and do not require any explicit labels.

Given a reference $R$, the encoder $Enc_r$ compresses domain information into a fixed-length vector; we call this a \textit{reference embedding}. This embedding is used as a query vector in the universal token layer $UTL$, which consists of a bank of \textit{token embedding} and an attention layer, where the token embedding are randomly initialized. We use the attention layer to learn the similarity between the reference embedding and each of the tokens. This produces a set of weights that represent the contribution of each token. The weighted sum of these token embedding, which we call \textit{domain embedding}, is used as the encoded latent code $z$ for generation. The bank of token embedding is shared across all training sequences. Note that this whole process can be understood as information bottleneck~\cite{tishby2000information}, which allows our model to learn a highly structured latent space.

\subsection{Training Objectives} \label{sec:objective}
\textbf{cVAE-GAN}:
\label{sec:gan}
When taking both source $S$ and reference $R$ samples as input, it is natural 
to use conditional VAE-GANs as a learning objective:~\cite{mirza2014conditional}:
\begin{equation}
\begin{split}
    \mathcal{L}_{GAN}=\mathbb{E}_{S,T \in p(S,T)}[log(D(S, T))] \\
     + \mathbb{E}_{S, T \in p(S,T), z \sim E_{att}(R)}[log(1-D(S, G(S, z)))]
\end{split}
\end{equation}
A reconstruction loss is also adopted between the output and the ground truth:
\begin{equation}
    \mathcal{L}_{rec}=\mathbb{E}_{S,T \sim p(S,T), z \sim E_{att}(R)} \left\| \mathbf{T}, G(S,z) \right\|
\end{equation}
Further, we encourage the latent distribution, encoded by $Enc_{r}$, to be close to a random Gaussian. 
This allows us to randomly sample a latent code as the reference embedding when there is no references $R$ at testing time.
\begin{equation}
    \mathcal{L}_{kl}=E_{R \sim p(R)} \mathcal{D}_{kl}Enc_r(R) \vert \vert \mathcal{N}(0, I)
\end{equation}

The full objective for conditional VAE-GAN is:

\begin{equation}
    \mathcal{L}_{GAN}^{VAE}= \mathop{\min}\limits_{G, Enc_{att}}\mathop{\max}\limits_{D} {\mathcal{L}_{GAN}
    + \lambda_{rec}\mathcal{L}_{rec} + \lambda_{kl}\mathcal{L}_{kl}}
\end{equation}

This process is shown in Figure \ref{fig:pipeline} (yellow lines).

\begin{figure*}[t]
\begin{center}
    \includegraphics[width=1.0\linewidth]{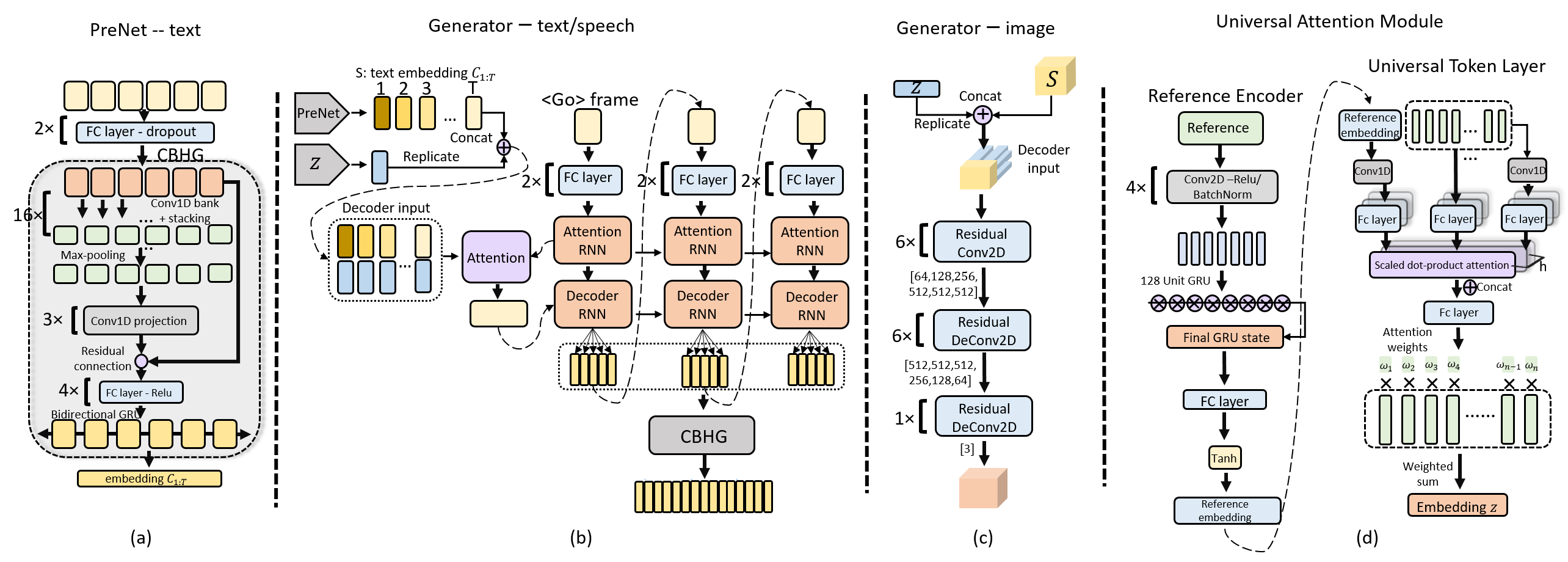}
\end{center}
   \caption{Diagrams of each module's architecture in M$^{3}$D-GAN.}
\label{fig:network-architecture}
\end{figure*}

\textbf{Latent Regression}: 
We find that, by training the model only on the conditional VAE-GAN objective, the synthesized samples tend to be very similar even with different references.
In other words, the latent codes output from $Enc_r$ are ignored by the model.
In this way, when training jointly with the whole model but without any constraints,
the universal attention module (includes $Enc_r$ and $UTL$) is hard to optimize.
This issue has been pointed out in \cite{VQ-VAE}, and named ``posterior collapse''.

We assume that if using $Enc_r$ to encode a synthesized sample $\hat{T}$, then the 
output latent code $\hat{z}$ should be highly correlated with the one used to synthesize itself, i.e. $ (S, z) \to \hat{T} \to \hat{z}$, $\|z, \hat{z}\|$.
Thus, a solution would be to directly apply regression on the latent code $z$ 
that is encoded from reference $R$, i.e. $R \to z, (S, z) \to \hat{T} \to \hat{z}$,
$\|z \sim Enc_r(R), \hat{z}\|$.

However, if $Enc_r$ collapses, i.e., it encodes any input to the same or similar latent codes, the reconstructed code will still be the same and $Enc_r$ will not be trained in an optimal fashion.
Therefore, rather than using a latent code that is encoded from the reference sample, i.e. $z \sim Enc_r(R)$, we start the latent regression process via a randomly drawn latent code $z \sim \mathcal{N}(0, I)$ and attempt to recover it, i.e. $z \sim \mathcal{N}(0, I), (S, z) \to \hat{T} \to \hat{z}$, $\|z \sim \mathcal{N}(0, I), \hat{z}\|$. 
We denote the randomly sampled latent code as $z_s$,
and the one encoded from $R$ as $z_r$.
The samples generated by $z_s$ as $\hat{T}_{sam}$, the ones generated by $z_r$ as $\hat{T}_{enc}$. Thus the latent regression loss $\mathcal{L}_{lat}$ is:
\begin{equation}
    \mathcal{L}_{lat} = \mathbb{E}_{S \sim p(S), z_s \sim \mathcal{N}(0, I)} \| z_s-Enc_r(G(S,z_s))\|_1
\end{equation}
It is natural to also use the discriminator loss on $\hat{T}_{sam}$. The full loss can be written as:
\begin{equation}
    \mathcal{L}_{GAN}^{lat}=\mathop{\min}\limits_{G, Enc_{att}}\mathop{\max}\limits_{D}{\mathcal{L}_{GAN} + \lambda_{lat}\mathcal{L}_{lat}}
\end{equation}
This process is shown in Figure \ref{fig:pipeline} (green lines).

\textbf{Distance Regularization}
To further enforce the model to produce effective latent codes and prevent the model from collapsing, we propose a regularization on the generator that directly penalizes the mode collapse behavior. 
Our regularization is inspired by \cite{distance-GAN} which showed that the distance between a pair of samples should be highly related before and after translation. Thus, we design our regularization as:
\begin{equation} \label{eq:distance}
    \begin{split}
        \mathop{\max}\limits_{G}\mathcal{L}_{dist}=\mathbb{E}_{z_s,z_r}\left[  \frac{\left\|G(S, z_s)-G(S, z_r)) \right\| }{\left\| z_s-z_r \right\|} \right ]
    \end{split}
\end{equation}
where $\left \| \cdot  \right \|$ indicates a norm.
This regularization will penalize when the generator collapses into few modes, and thus force the generator not to ignore the latent code and produce diverse outputs. 

Our full objective function can be written as:
\begin{equation}
    \mathcal{L}_{all}=\mathop{min}\limits_{G, E_{att}}\mathop{max}\limits_D \lambda_{1}\mathcal{L}_{GAN}^{VAE} + \lambda_{2} \mathcal{L}_{GAN}^{lat} + \lambda_{3} \mathcal{L}_{dist}
\end{equation}

\section{Network Architecture} 
\textbf{Reference Encoder} $Enc_r$ $\&$ \textbf{universal token layer} $UTL$: The input to $Enc_r$ is first passed through a stack of four 2-D convolutional layers with [64, 64, 128, 128] channels. To adapt to sequential input, we make the output tensor to preserve the time resolution and feed it into a single-layer 128-unit GRU. The last GRU state serves as the reference embedding, which is then fed to the universal token layer.
We use 10 tokens in our experiments, which we found sufficient to represent a small but rich variety of domain information in the training data. We use multihead attention~\cite{vaswani2017attention} to compute the attention weights. It uses a softmax activation to output a set of combination weights over the tokens; the resulting weighted combination of these tokens is the domain embedding.

\textbf{modality subnets (Prenet)} $\mathbf{M}_{in}$: We process speech data via mel-spectrograms which enables our model to take both images and speech via a two-layer fully 2D-convolutional network with dimension [32, 32]. For text input,
we feed a sequence of 128-D character level embeddings into two fully-connected layers with [256, 128] units. 
The output is fed into a CBHG unit~\cite{Tacotron} which has a Conv1D bank with 16 layers, and each layer has 128 units. After the residual connection, there are four 128 unit fully-connected layers with ReLU. The final Bidirectional GRU has 128 cells.

\textbf{modality subnets (Generator)} $\mathbf{M}_{out}$:
As speech and text are both sequential signals, we design the same generator architecture for them. It takes combined $S$ and $z$ as input. To match the dimension, we first replicate the single $z$ T times, and concatenat it with the input source sequence. The Attention RNN consists of 2-layer residual GRUs with 256 cells. The Decoder RNN has a 256 cell 1-layer GRU. When synthesizing speech, this generator directly predicts the mel-spectrogram. We use Griffin-Lim~\cite{griffin-lim} as a vocoder. To process image data, six residual conv2D layers compress the input into a low dimension representation, and then six residual deconv2D layers are used for decoding. Finally, a deconv2D layer outputs 3D RGB images.

Figure \ref{fig:network-architecture} shows detailed diagrams for each module. We train our M$^{3}$D-GAN from scratch using the Adam-optimizer. For the task of image $\to$ image translation, we train our model for 30 epochs with a batch size of 1. For other tasks, we train our model with a batch size of 32. For text $\to$ image, and image $\to$ text, we train our model for 300 epochs. Text $\to$ speech and text $\to$ text are trained for 200k steps. Details can be found in the supplementary material.

%-------------------------------------------------------------------------
\section{Experiments}
We test the proposed approach on six synthesis tasks and compare (quantitatively and qualitatively) to baseline methods in each case. The datasets we used, and baseline methods we compared against, are listed in Table~\ref{tab:tasks}.

\begin{table}[]
\caption{The datasets and baseline methods for each task.}
\vspace{-10pt}
\begin{center}
\begin{tabular}{p{4.4cm}p{3.2cm}}
\toprule
Dataset & Baseline Methods   \\ \hline \hline
   \makecell[l]{\textbf{Image-to-Image} \\ Edges$\to$photos~\cite{edges2shoes,edges2photo} \\ Outdoor day$\to$Night images~\cite{day2night}}  & \makecell[l]{Pix2Pix~\cite{pix2pix} \\ Bicycle-GAN~\cite{Bicycle-GAN} \\ cVAE-GAN~\cite{bao2017cvae}} \\ \hline
   \makecell[l]{\textbf{Text-to-Image} \\ CUB-200-2011 \cite{CUB_200_2011}}   &  \makecell[l]{StackGAN \cite{stackgan}\\ DA-GAN \cite{DA-GAN}}    \\  \hline
  \makecell[l]{\textbf{Image-to-Text} \\ MSCOCO \cite{MSCOCO}}  & CNN-RNN-coco \cite{show&tell} \\  \hline
  \makecell[l]{\textbf{Text-to-Speech} \\ EMT-4 
22,377 American \\ English audio-text samples.} & \makecell[l]{Tacotron2~\cite{Tacotron} \\
    Tacotron-GST~\cite{DBLP:conf/icml/Wang-GST_Tacotron}}  \\ \hline
\makecell[l]{\textbf{Speech-to-Text} \\
    LibriSpeech} & DeepSpeech2 \cite{DBLP:DeepSpeech} \\ \hline
\makecell[l]{\textbf{Text-to-Text} \\
    WMT'14 (En-Fr)} & JointRNN \cite{machien-translation} \\
\bottomrule
\end{tabular}
\label{tab:tasks}
\end{center}
\vspace{-8mm}
\end{table}

\subsection{Qualitative Evaluation} \label{sec:quality}
\textbf{Image $\to$ Image}: 
We test our model under two scenarios shown in Figure \ref{fig:pipeline}. The first is to provide a reference image $R$, i.e., $(S,R) \to \hat{T}$ and synthesize samples that exhibit the content of $S$ and the style of the reference sample. Figure \ref{fig:paris} shows the result: When given different references (first row), the synthesized samples preserve the content from the source image (the Eiffel Tower), and show different styles taken from references. 
The second scenario is to provide a random noise vector $z$ and synthesize a new sample by taking $S$ combined with $z$, i.e. $(S,z) \to \hat{T}$. Figure \ref{fig:paris} (bottom row) shows the results: When adding randomly sampled noise vector $z$, we obtain images of the same content with different styles.

\begin{figure}[t]
\begin{center}
    \includegraphics[width=0.8\linewidth]{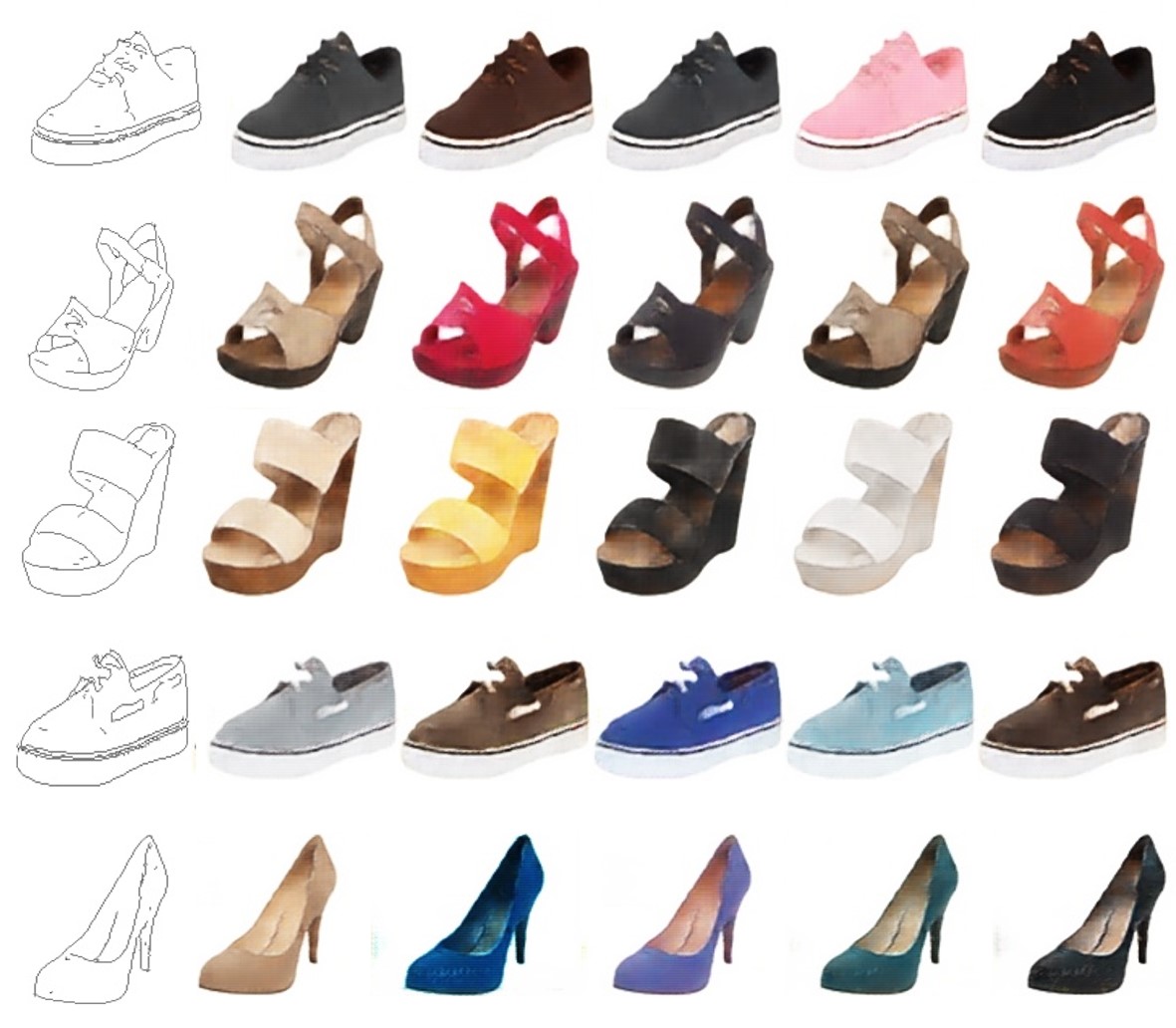}
\end{center}
   \caption{Edges $\to$ shoes generation by combining with randomly sampled noise vectors $z$ at testing time. For each row, the first column is the source sketch image.}
\label{fig:shoes_noise}
\end{figure}

\begin{figure*}[h]
\begin{center}
    \includegraphics[width=1\linewidth]{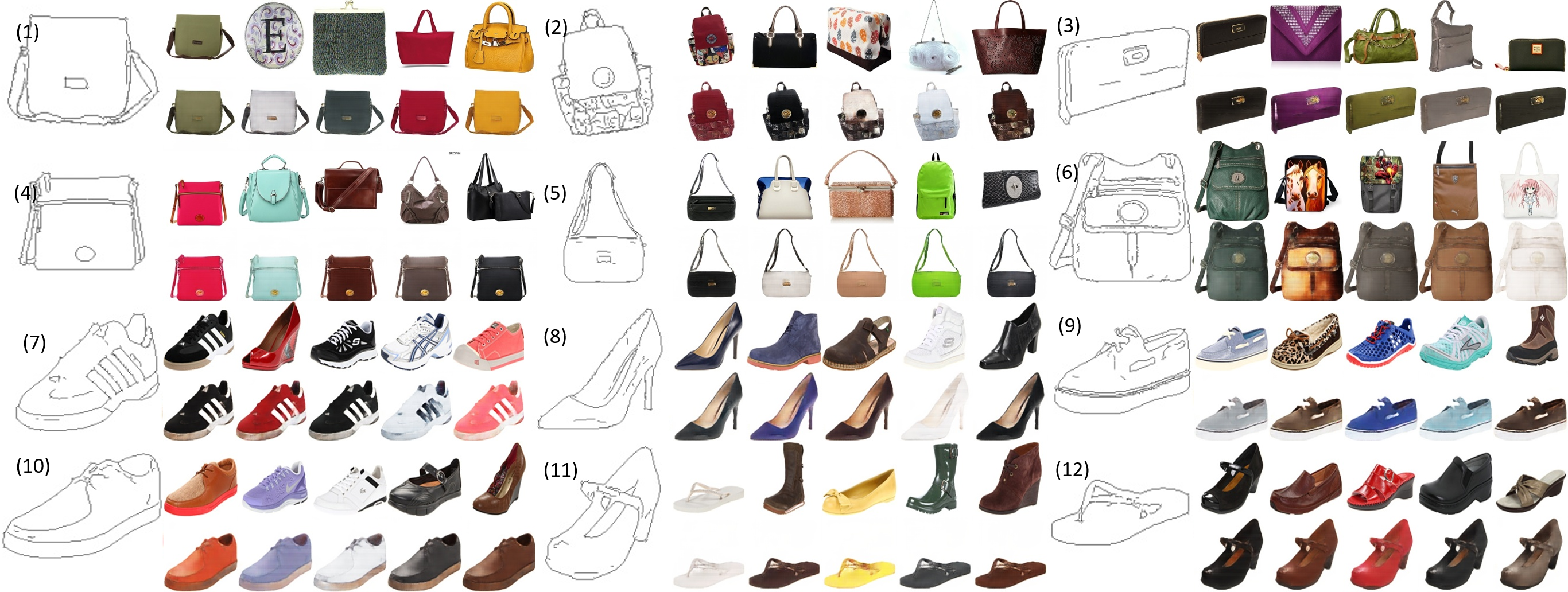}
\end{center}
   \caption{Explicit controlling for Image $\to$ Image. In each set, the first row 
   is the reference image $r$, and the second row is synthesized images corresponding 
   to the references domain. Where in each set, row 1, column 1 is the ground truth image, thus the image in row2, column1 can be considered as the reconstruction results.}
\label{fig:img2img}
\end{figure*}

More results are shown in Figure~\ref{fig:shoes_noise} and Figure~\ref{fig:img2img}. Figure~\ref{fig:shoes_noise} shows diverse images are produced when we provide a random noise vector to our model, which suggests that our universal attention module can alleviate the mode collapse problem. Figure~\ref{fig:img2img}, on the other hand shows our model correctly capturing domain information from any given references. For example, when generating shoes from sketches (set (7)), we can see that the sneaker's white stripes are correctly generated, while the shoes body colors are varied. Also, when given a reference that has a grey strip pattern (the fourth reference in this set), our model synthesizes a white body color with grey stripes. In another set (set (8)), the outside of the high heel shoes are correctly changed to the color of the reference, while the inside material remains the same. These results suggests that the model does take the reference's domain information, while successfully preserving the content information from the source image.

\begin{figure}[t]
\begin{center}
    \includegraphics[width=1\linewidth]{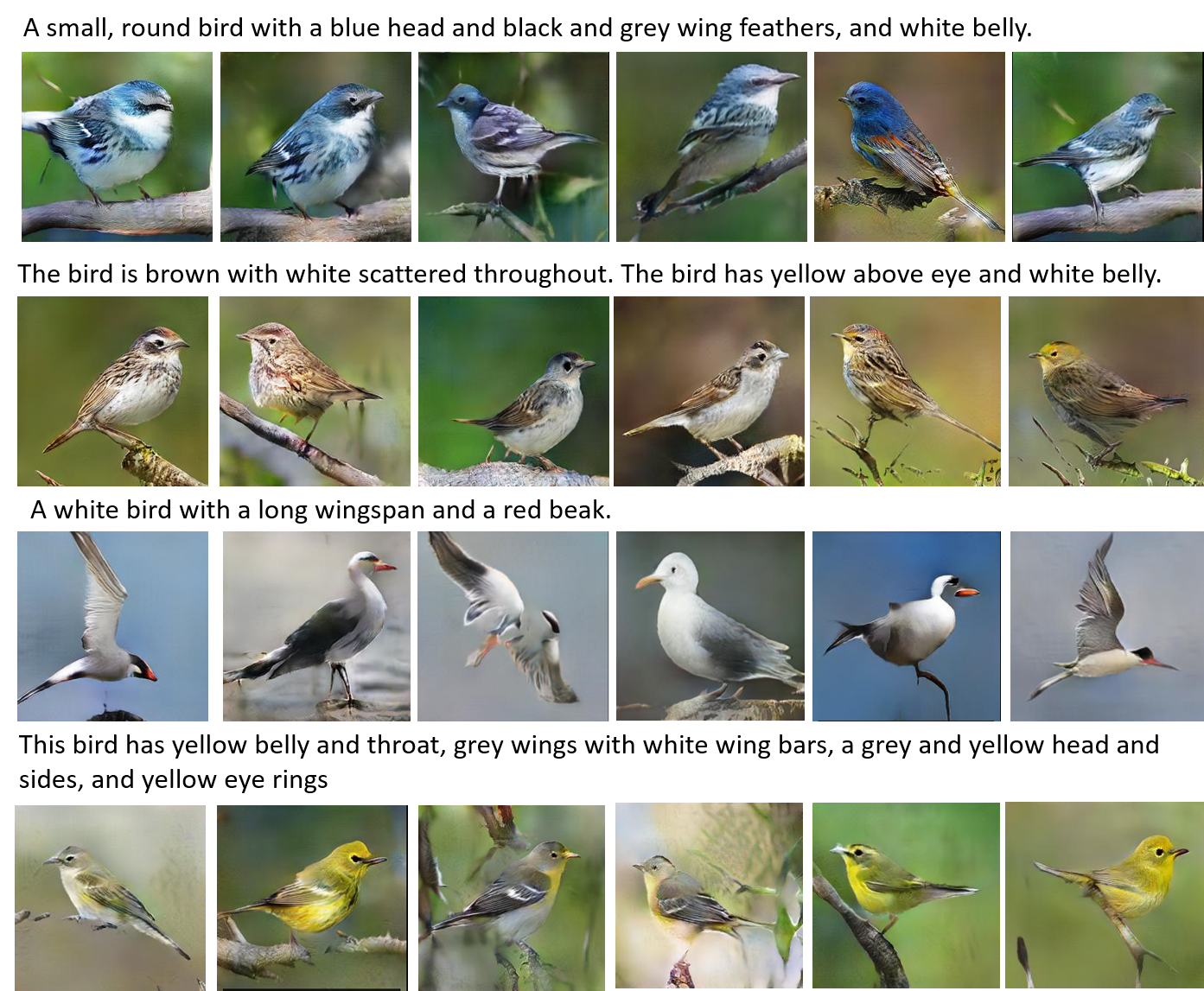}
\end{center}
   \caption{Text $\to$ Images. For each sentence, we randomly sampled 6 noise vectors for generating.}
\label{fig:text2img}
\end{figure}

\textbf{Text $\to$ Image}: We demonstrate our model's ability to perform cross-modal translation. We use the CUB-200-2011 dataset~\cite{CUB_200_2011}, where each image is paired with 10 sentences describing a bird in the image. We resize input images to have a 256 $\times$ 256 pixel resolution for 
better visual quality. During training, we randomly sample a sentence and pair it with an image. At inference, we provide a random noise vector along with the input sentence.
Figure~\ref{fig:text2img} shows the results. Given a sentence description, our model produces images that correctly preserve certain characteristics of a bird described by the text (e.g., bird shape) while showing diversity in other attributes (e.g., pose). 

\textbf{Text $\to$ Speech}: We test our model's ability to generate stylized speech from text inputs. In conventional TTS systems, training on a dataset of (text, audio) pairs, the model is supposed to synthesize an audio clip from a given textual string. The audio needs to convey the correct content of the text, and possess characteristics of the human voice. For adding styles to TTS, at inference, we combine an additional reference audio with 
a textual string and feed them into the system, where the text provides content, and the reference audio provides the desired style. For example, when synthesizing a sentence `The cat is laying on the table'. Given a reference audio signal which has happy emotional tone, the synthesized sample
should say the sentence in happy tone. Note that, the reference audio signal can contain any textual content.
To this end, the most challenging part is in disentangling the content and style information from the reference audio. We perform experiments to validate that our model can successfully capture domain (style) information  from the reference audio.

We compare our model to Tacotron-GST~\cite{DBLP:conf/icml/Wang-GST_Tacotron} in how well they model auditory styles. We use an in-house dataset, dubbed EMT-4, that consists of 22,377 American English audio-text samples, with a total of 24 hours. All the audio samples are read by a single speaker in four emotion categories: happy, sad, angry and neutral. For each text sample, there is only one audio sample paired with one of the four emotion styles.

% \dan{Do we mention EMT-4 before this in the paper? We might need to explain}. 
We randomly select 15 sentences and 4 references from 4 emotion categories. Each sentence is paired with four reference audio samples for synthesis, producing 60 audio samples. We put the audio results on our \href{https://researchdemopage.wixsite.com/cvpr-tts-demo/style-transfer} {demo} page. 
To see whether our model can capture domain information from references, we also conduct a ``content-style swap'' experiment where samples are synthesized by permuting text (content) and audio (style) from four (text, audio) pairs, one from each emotion category. We put the results on our \href{https://researchdemopage.wixsite.com/cvpr-tts-demo/style-and-content-swapping}{demo} page; each column has the same content with different styles, each row has the same style with different content. The results suggests that our model has successfully disentangled content and style components.

\begin{table*} 
\caption{Quantitative results and comparisons with the state-of-the-arts methods for different task .}
    \begin{subtable}[t]{.5\linewidth}
      \caption{Evaluation on Image $\to$ Image.}
      \centering
        \begin{tabular}[t]{rcc}
\toprule
Method & \begin{tabular}[c]{@{}l@{}}Realism (acc)\\ svhn $\to$ mnist\end{tabular}  & \begin{tabular}[c]{@{}l@{}}Diversity(LPIPS)\\ edges $\to$ shoes\end{tabular}   \\ \hline \hline
pix2pix~\cite{pix2pix}  & 89\% & .013$\pm$.000 \\
cVAE-GAN~\cite{bao2017cvae} & 86\% & .096$\pm$.001  \\
Bicycle-GAN~\cite{Bicycle-GAN} & 91\% & .110$\pm$.002 \\
\textbf{Ours($z$)} & \textbf{91\% } &  \textbf{.112$\pm$.001}        \\
\textbf{Ours($r$)} & -- &  \textbf{.115$\pm$.002} \\ \hline
Rand. Real Imgs. & 98\%  & .262$\pm$.007 \\ 
\bottomrule
\end{tabular}
\label{tab:image-to-image}
    \end{subtable}%
    \begin{subtable}[t]{.5\linewidth}
      \centering
        \caption{Evaluation on tasks of Image$\to$Text \& Text$\to$Text.}
        \begin{tabular}[t]{c|cc}
\toprule
& \multicolumn{2}{c}{\begin{tabular}[c]{@{}c@{}}Image$\to$Text
(Image Caption)\end{tabular}}  \\ %\hline
Method  & BLEU-1   & BLEU-4     \\ \hline \hline
CNN+RNN-coco~\cite{show&tell}  & 66.7  & 23.8   \\ %\cline{1-3}
Ours &  65.2  & 23.8 \\     
& \multicolumn{2}{c}{\begin{tabular}[c]{@{}c@{}} Text$\to$Text (Machine Trans.)\end{tabular}} \\ 
JointRNN (En-Fr)~\cite{machien-translation}     &  -- & 28.4   \\ %\cline{1-3}
Ours   &  -- & 22.2 \\     
\bottomrule
\\
\end{tabular}
\label{tab:image-to-text}
    \end{subtable} 
    \begin{subtable}[t]{.4\linewidth}
      \caption{Evaluation Text $\to$ Image synthesis task.}
      \centering
        \begin{tabular}[t]{ccc}
\toprule
Method    &    Inception   &    $\#$miss   \\ \hline \hline
StackGAN~\cite{stackgan}  & 3.7$\pm$0.4 & 36.0   \\
DA-GAN~\cite{DA-GAN}       &  5.6$\pm$0.4         & 19.0                  \\
Ours & \textbf{5.2$\pm$0.4} &  \textbf{16.0} \\
\bottomrule
\end{tabular}
\label{tab:text-to-image}
    \end{subtable}%
    \begin{subtable}[t]{.3\linewidth}
      \centering
        \caption{Evaluation on Text$\to$Speech (TTS).}
        \begin{tabular}[t]{c|cc}
\toprule
Method & WER & Acc \\ \hline \hline
Tac.~\cite{Tacotron}     & 10.6     &   68\%  \\ %\cline{1-4}
GST~\cite{DBLP:conf/icml/Wang-GST_Tacotron}   & \textbf{10.2}       &  77\%       \\
Ours & \textbf{10.2}  & \textbf{80\%}  \\
\bottomrule
\end{tabular}
\label{tab:text-to-speech}
    \end{subtable} 
    \begin{subtable}[t]{.3\linewidth}
      \centering
        \caption{Evaluation on Speech Recognition.}
        \begin{tabular}[t]{c|cc}
\toprule
Metric &           WER                                                                  \\ \hline \hline
DeepSpeech2~\cite{DBLP:DeepSpeech}    & 5.15  \\ %\cline{1-4}
Ours      & 7.3\\
\bottomrule
\end{tabular}
\label{tab:speech-to-text}
    \end{subtable} 

\end{table*}

\subsection{Quantitative Evaluation} \label{sec:quantity}
\textbf{Image $\to$ Image}:
We evaluate our model in terms of \textbf{realism} and \textbf{diversity} when synthesizing images from images. We train a classifier on the MNIST dataset~\cite{mnist} and employ it on the translated samples from SVHN~\cite{svhn}. Table~\ref{tab:image-to-image} shows that our model achieves comparable results in realism with the state-of-the-art. To evaluate diversity, we compute the diversity-score using the LPIPS metric \cite{LPIPS}. Taking 100 shoe sketches, we randomly generate 2000 images using each model by adding 
randomly sampled latent codes ($z$). The average distance between the 2000 samples for each method are reported in Table \ref{tab:image-to-image}. Our model (Ours($z$)) produces the highest diversity score. While pix2pix \cite{pix2pix} has the lowest diversity, potentially due to the mode collapse issue in conditional GANs. We also generate samples by adding references. The diversity score is further improved (see Ours($R$)), illustrating that explicitly controlling the domain variable leads to better results.

\textbf{Text $\to$ Image}:
We use two metrics: the Inception Score~\cite{improved-techniques-for-training-gan} and the number of missing modes (denoted as $\#$ miss). To compute the inception score, we finetune VGG-19 on CUB-200-2011~\cite{CUB_200_2011}. Table~\ref{tab:text-to-image} shows the results. Comparing with DA-GAN~\cite{DA-GAN} which requires labels for training, our model produces comparable results even without using any label.
For a more rigorous validation, we also adopt the missing mode metric ($\#$ miss) that represents the classifier reported number of missing modes. Comparing with the other two methods, our model misses the least number of modes, which further suggests the superiority of our model in alleviating the mode collapse issue.

\textbf{Image $\to$ Text (image captioning):}
We compare our model with CNN-RNN-COCO, a state-of-the-art image caption method based on \cite{show&tell}. Table \ref{tab:image-to-text} shows the results on MS-COCO; our model achieves BLEU-4 in 0.238 for both. Comparing with the model that specifically designed for this task, our model produces comparable results.

\textbf{Text $\to$ Speech (TTS)}
We evaluate two factors in the TTS task, i.e. fidelity (the synthesized speech should contain the desired content in a clearly audible form) and domain transfer accuracy (when transferring across domains, the synthesized speech should correctly correspond to the reference's domain). We compare with two state-of-the-art stylized TTS models, Tacotron and GST, where Tacotron is an RNN-CNN auto-regressive model trained only on reconstruction loss, and GST improves on Tacotron by incorporating the global style token layers.

To validate fidelity, we assess the performance of synthesized samples in a speech recognition task. We use a pre-trained ASR model based on WaveNet \cite{WaveNet} to compute the Word Error Rate (WER) for the samples synthesized by each model. Table \ref{tab:text-to-speech} shows our model performing comparably with the state-of-the-art approaches. 
% Note that WER measures only the correctness of verbal content, not its auditory style. The results suggests that all the methods we have comparable perform reasonably well in controlling the verbal content in TTS. 
As our domains (emotions) are all categorical, we evaluate the performance in domain transfer by means of classification. To this end, we train a classifier on EMT-4, which shows a 98$\%$ accuracy. We then select 1000 samples synthesized from the test set of EMT-4. Table \ref{tab:text-to-speech} shows that our model performs the best in terms of the domain transfer accuracy.

\textbf{Speech $\to$ Text (speech recognition) and Text $\to$ Text (machine translation)}
We test our model for the speech recognition task on LibriSpeech corpus constructed from audio books \cite{LibriSpeech}. For text-to-text, we test our model on the WMT'14 English-French (En-Fr) dataset. We remove sentences longer than 175 words, resulting in 35.5M sentence pairs for training. The source and target vocabulary is based on 40k BPE types. We evaluate with tokenized BLEU on the corpus-level. As seen in Table \ref{tab:image-to-text} and Table \ref{tab:speech-to-text}, comparing with the state-of-the-art speech recognition method, i.e. DeepSpeech \cite{DBLP:DeepSpeech} and machine translation method, our model did not show outstanding results. We suspect this is because when translating text to text, the reference (text) will be processed as a single domain embedding, which is a global representation, and thus lose essential sequential information. Even though our generator decodes information by time step and works well for other modalities (e.g., speech), we suspect that our text encoding method could be suboptimal. Future work may focus on designing a more robust network architecture for the text modality.

\subsection{Ablation Study} \label{sec:ablation_study}
%\vspace{-3mm}

\begin{table}[t]
      \caption{Evaluation for each component of M$^{3}$D-GAN.}
      \vspace{-3mm}
      \centering
        \begin{tabular}[t]{rcc}
\toprule
Model & \begin{tabular}[c]{@{}l@{}}Realism (acc)\\ svhn $\to$ mnist\end{tabular}  & \begin{tabular}[c]{@{}l@{}}Diversity(LPIPS)\\ edges $\to$ shoes\end{tabular}   \\ \hline \hline
cVAE-GAN~\cite{bao2017cvae} & 86\% & .096$\pm$.001  \\ \hline
$\mathcal{L}_{GAN}^{VAE}$  & 87\% & .098$\pm$.001 \\
$\mathcal{L}_{GAN}^{VAE}$ + $\mathcal{L}_{GAN}^{lat}$ & 90\% & .113$\pm$.000  \\
$\mathcal{L}_{all}$ w$\setminus$o Att & 89\%  &  .111$\pm$.001        \\
$\mathcal{L}_{all}$ & \textbf{91 \%} &  \textbf{.115$\pm$.002} \\ \hline
Rand. Real Imgs. & 98\%  & .262$\pm$.007 \\ 
\bottomrule
\end{tabular}
\label{tab:ablation_study}
\vspace{-0.5cm}
\end{table}%

We conduct an ablation study on the image-to-image task to analyze each component in our model. We again use the classification accuracy and diversity score as the metric. The results are shown in Table \ref{tab:ablation_study}. 
\begin{itemize}
    \item cVAE-GAN~\cite{bao2017cvae} vs. $\mathcal{L}^{VAE}_{GAN}$: When training our model with the conventional conditional VAE-GANs objective~\cite{bao2017cvae}, we see improvements in terms of both the realism and diversity. It shows the contribution of our attention module in producing a highly structured latent space, which helps the model generate more realistic and diverse results.
    \item $\mathcal{L}^{VAE}_{GAN}$ vs. $\mathcal{L}^{VAE}_{GAN} + \mathcal{L}^{lat}_{GAN}$: When we add the latent regression, the performance is further improved.
    \item $\mathcal{L}^{VAE}_{GAN} + \mathcal{L}^{lat}_{GAN}$ vs. $\mathcal{L}_{all}$: Comparing with our model with all the proposed losses, the realism of the $\mathcal{L}^{VAE}_{GAN} + \mathcal{L}^{lat}_{GAN}$ baseline is slightly impaired, and the diversity is also lower. It shows that, the distance regularization $L_{dist}$ plays an important role in controlling the learned latent code.
    \item $\mathcal{L}_{all}$ w$\setminus$o Att vs. $\mathcal{L}_{all}$: When discarding the attention module, both realism and diversity are lower than the proposed model. It illustrates that the attention mechanism does help in producing better results.
\end{itemize}

%\vspace{-3mm}
%------------------------------------------------------------------------
\section{Conclusion}
%\vspace{-3mm}
We present M$^3$D-GAN for cross-modal cross-domain translation, which consists of modality-specific subnets with a universal attention module that learns to encode modality/domain information in a unified way. We show how the same architecture can be applied to a wide variety of tasks including text-to-image, image-to-text, image-to-image, text-to-speech, speech-to-text and text-to-text translation. The universal attention module we propose can learn a highly structured latent space by means of information bottleneck~\cite{tishby2000information}, which allows for explicit control of specific attributes. Leveraging the same architecture for different problems is advantageous as it allows us to develop shared representations in the abstracted unified space. We conduct a comprehensive set of experiments, both qualitative and quantitative, to show that our model achieves strong results across most of the tasks, especially for speech and image. 

\balance{}

{%\small
\bibliographystyle{ieee}
\bibliography{egbib}

\begin{thebibliography}{10}\itemsep=-1pt

\bibitem{DBLP:DeepSpeech}
D.~Amodei, S.~Ananthanarayanan, R.~Anubhai, J.~Bai, E.~Battenberg, C.~Case,
  J.~Casper, B.~Catanzaro, Q.~Cheng, G.~Chen, J.~Chen, J.~Chen, Z.~Chen,
  M.~Chrzanowski, A.~Coates, G.~Diamos, K.~Ding, N.~Du, E.~Elsen, J.~Engel,
  W.~Fang, L.~Fan, C.~Fougner, L.~Gao, C.~Gong, A.~Hannun, T.~Han, L.~V.
  Johannes, B.~Jiang, C.~Ju, B.~Jun, P.~LeGresley, L.~Lin, J.~Liu, Y.~Liu,
  W.~Li, X.~Li, D.~Ma, S.~Narang, A.~Ng, S.~Ozair, Y.~Peng, R.~Prenger,
  S.~Qian, Z.~Quan, J.~Raiman, V.~Rao, S.~Satheesh, D.~Seetapun, S.~Sengupta,
  K.~Srinet, A.~Sriram, H.~Tang, L.~Tang, C.~Wang, J.~Wang, K.~Wang, Y.~Wang,
  Z.~Wang, Z.~Wang, S.~Wu, L.~Wei, B.~Xiao, W.~Xie, Y.~Xie, D.~Yogatama,
  B.~Yuan, J.~Zhan, and Z.~Zhu.
\newblock Deep speech 2: End-to-end speech recognition in english and mandarin.
\newblock In {\em Proceedings of the 33rd International Conference on
  International Conference on Machine Learning - Volume 48}, ICML'16, 2016.

\bibitem{machien-translation}
D.~Bahdanau, K.~Cho, and Y.~Bengio.
\newblock Neural machine translation by jointly learning to align and
  translate.
\newblock {\em CoRR}, abs/1409.0473, 2014.

\bibitem{bao2017cvae}
J.~{Bao}, D.~{Chen}, F.~{Wen}, H.~{Li}, and G.~{Hua}.
\newblock Cvae-gan: Fine-grained image generation through asymmetric training.
\newblock In {\em 2017 IEEE International Conference on Computer Vision
  (ICCV)}, 2017.

\bibitem{distance-GAN}
S.~Benaim and L.~Wolf.
\newblock One-sided unsupervised domain mapping.
\newblock In {\em Proceedings of the 31st International Conference on Neural
  Information Processing Systems}, NIPS'17, 2017.

\bibitem{MSCOCO}
X.~Chen, H.~Fang, T.~Lin, R.~Vedantam, S.~Gupta, P.~Doll{\'{a}}r, and C.~L.
  Zitnick.
\newblock Microsoft {COCO} captions: Data collection and evaluation server.
\newblock {\em CoRR}, 2015.

\bibitem{choi2017stargan}
Y.~Choi, M.~Choi, M.~Kim, J.-W. Ha, S.~Kim, and J.~Choo.
\newblock Stargan: Unified generative adversarial networks for multi-domain
  image-to-image translation.
\newblock In {\em The IEEE Conference on Computer Vision and Pattern
  Recognition (CVPR)}, June 2018.

\bibitem{attribute}
A.~Farhadi, I.~Endres, D.~Hoiem, and D.~Forsyth.
\newblock Describing objects by their attributes.
\newblock In {\em CVPR}, 2009.

\bibitem{giard1999auditory}
M.~H. Giard and F.~Peronnet.
\newblock Auditory-visual integration during multimodal object recognition in
  humans: a behavioral and electrophysiological study.
\newblock {\em Journal of cognitive neuroscience}, 11(5):473--490, 1999.

\bibitem{GANs}
I.~Goodfellow, J.~Pouget-Abadie, M.~Mirza, B.~Xu, D.~Warde-Farley, S.~Ozair,
  A.~Courville, and Y.~Bengio.
\newblock Generative adversarial nets.
\newblock In {\em NIPS}. 2014.

\bibitem{griffin-lim}
D.~Griffin and J.~Lim.
\newblock Signal estimation from modified short-time fourier transform.
\newblock {\em IEEE Transactions on Acoustics, Speech, and Signal Processing},
  32(2):236--243, April 1984.

\bibitem{pix2pix}
P.~{Isola}, J.~{Zhu}, T.~{Zhou}, and A.~A. {Efros}.
\newblock Image-to-image translation with conditional adversarial networks.
\newblock In {\em 2017 IEEE Conference on Computer Vision and Pattern
  Recognition (CVPR)}, 2017.

\bibitem{kohler2002hearing}
E.~Kohler, C.~Keysers, M.~A. Umilta, L.~Fogassi, V.~Gallese, and G.~Rizzolatti.
\newblock Hearing sounds, understanding actions: action representation in
  mirror neurons.
\newblock {\em Science}, 297(5582):846--848, 2002.

\bibitem{day2night}
P.-Y. Laffont, Z.~Ren, X.~Tao, C.~Qian, and J.~Hays.
\newblock Transient attributes for high-level understanding and editing of
  outdoor scenes.
\newblock {\em SIGGRAPH}, 33(4), 2014.

\bibitem{mnist}
Y.~LeCun and C.~Cortes.
\newblock {MNIST} handwritten digit database.
\newblock 2010.

\bibitem{DA-GAN}
S.~Ma, J.~Fu, C.~Wen~Chen, and T.~Mei.
\newblock Da-gan: Instance-level image translation by deep attention generative
  adversarial networks.
\newblock In {\em The IEEE Conference on Computer Vision and Pattern
  Recognition (CVPR)}, June 2018.

\bibitem{ma-iclr19}
S.~Ma, D.~Mcduff, and Y.~Song.
\newblock A generative adversarial network for style modeling in a
  text-to-speech system.
\newblock In {\em International Conference on Learning Representations}, 2019.

\bibitem{mirza2014conditional}
M.~Mirza and S.~Osindero.
\newblock Conditional generative adversarial nets.
\newblock {\em arXiv preprint arXiv:1411.1784}, 2014.

\bibitem{svhn}
Y.~Netzer, T.~Wang, A.~Coates, A.~Bissacco, B.~Wu, and A.~Y. Ng.
\newblock Reading digits in natural images with unsupervised feature learning.
\newblock In {\em NIPS Workshop on Deep Learning and Unsupervised Feature
  Learning 2011}, 2011.

\bibitem{odena2017conditional}
A.~Odena, C.~Olah, and J.~Shlens.
\newblock Conditional image synthesis with auxiliary classifier gans.
\newblock In {\em ICML}, 2017.

\bibitem{LibriSpeech}
V.~Panayotov, G.~Chen, D.~Povey, and S.~Khudanpur.
\newblock Librispeech: An {ASR} corpus based on public domain audio books.
\newblock In {\em ICASSP}. {IEEE}, apr 2015.

\bibitem{pietrini2004beyond}
P.~Pietrini, M.~L. Furey, E.~Ricciardi, M.~I. Gobbini, W.-H.~C. Wu, L.~Cohen,
  M.~Guazzelli, and J.~V. Haxby.
\newblock Beyond sensory images: Object-based representation in the human
  ventral pathway.
\newblock {\em Proceedings of the National Academy of Sciences},
  101(15):5658--5663, 2004.

\bibitem{text2image_reed}
S.~Reed, Z.~Akata, X.~Yan, L.~Logeswaran, B.~Schiele, and H.~Lee.
\newblock Generative adversarial text to image synthesis.
\newblock In {\em Proceedings of the 33rd International Conference on
  International Conference on Machine Learning - Volume 48}, ICML'16, 2016.

\bibitem{Tacotron}
R.~J. Skerry{-}Ryan, E.~Battenberg, Y.~Xiao, Y.~Wang, D.~Stanton, J.~Shor,
  R.~J. Weiss, R.~Clark, and R.~A. Saurous.
\newblock Towards end-to-end prosody transfer for expressive speech synthesis
  with tacotron.
\newblock {\em CoRR}, abs/1803.09047, 2018.

\bibitem{cross-domain-image-generation}
Y.~Taigman, A.~Polyak, and L.~Wolf.
\newblock Unsupervised cross-domain image generation.
\newblock In {\em ICLR}, 2017.

\bibitem{tishby2000information}
N.~Tishby, F.~C. Pereira, and W.~Bialek.
\newblock The information bottleneck method.
\newblock {\em arXiv preprint physics/0004057}, 2000.

\bibitem{WaveNet}
A.~van~den Oord, S.~Dieleman, H.~Zen, K.~Simonyan, O.~Vinyals, A.~Graves,
  N.~Kalchbrenner, A.~Senior, and K.~Kavukcuoglu.
\newblock Wavenet: A generative model for raw audio.
\newblock In {\em Arxiv}, 2016.

\bibitem{VQ-VAE}
A.~van~den Oord, O.~Vinyals, and k.~kavukcuoglu.
\newblock Neural discrete representation learning.
\newblock In {\em Advances in Neural Information Processing Systems 30}. 2017.

\bibitem{vaswani2017attention}
A.~Vaswani, N.~Shazeer, N.~Parmar, J.~Uszkoreit, L.~Jones, A.~N. Gomez,
  {\L}.~Kaiser, and I.~Polosukhin.
\newblock Attention is all you need.
\newblock In {\em NIPS}, 2017.

\bibitem{show&tell}
O.~Vinyals, A.~Toshev, S.~Bengio, and D.~Erhan.
\newblock Show and tell: A neural image caption generator.
\newblock In {\em CVPR}, 2015.

\bibitem{CUB_200_2011}
C.~Wah, S.~Branson, P.~Welinder, P.~Perona, and S.~Belongie.
\newblock {The Caltech-UCSD Birds-200-2011 Dataset}.
\newblock Technical Report CNS-TR-2011-001, California Institute of Technology,
  2011.

\bibitem{DBLP:conf/icml/Wang-GST_Tacotron}
Y.~Wang, D.~Stanton, Y.~Zhang, R.~Ryan, E.~Battenberg, J.~Shor, Y.~Xiao,
  Y.~Jia, F.~Ren, and R.~A. Saurous.
\newblock Style tokens: Unsupervised style modeling, control and transfer in
  end-to-end speech synthesis.
\newblock In {\em ICML}, 2018.

\bibitem{xiong2018microsoft}
W.~Xiong, L.~Wu, F.~Alleva, J.~Droppo, X.~Huang, and A.~Stolcke.
\newblock The microsoft 2017 conversational speech recognition system.
\newblock In {\em ICASSP}, 2018.

\bibitem{edges2photo}
A.~Yu and K.~Grauman.
\newblock Fine-grained visual comparisons with local learning.
\newblock In {\em CVPR}, Jun 2014.

\bibitem{stackgan}
H.~{Zhang}, T.~{Xu}, H.~{Li}, S.~{Zhang}, X.~{Wang}, X.~{Huang}, and
  D.~{Metaxas}.
\newblock Stackgan: Text to photo-realistic image synthesis with stacked
  generative adversarial networks.
\newblock In {\em 2017 IEEE International Conference on Computer Vision
  (ICCV)}, 2017.

\bibitem{LPIPS}
R.~Zhang, P.~Isola, A.~A. Efros, E.~Shechtman, and O.~Wang.
\newblock The unreasonable effectiveness of deep features as a perceptual
  metric.
\newblock In {\em The IEEE Conference on Computer Vision and Pattern
  Recognition (CVPR)}, 2018.

\bibitem{zhang2017very}
Y.~Zhang, W.~Chan, and N.~Jaitly.
\newblock Very deep convolutional networks for end-to-end speech recognition.
\newblock In {\em ICASSP}, 2017.

\bibitem{cycleGAN}
J.~{Zhu}, T.~{Park}, P.~{Isola}, and A.~A. {Efros}.
\newblock Unpaired image-to-image translation using cycle-consistent
  adversarial networks.
\newblock In {\em 2017 IEEE International Conference on Computer Vision
  (ICCV)}, 2017.

\bibitem{edges2shoes}
J.-Y. Zhu, P.~Kr{\"a}henb{\"u}hl, E.~Shechtman, and A.~A. Efros.
\newblock Generative visual manipulation on the natural image manifold.
\newblock In {\em ECCV}, 2016.

\bibitem{Bicycle-GAN}
J.-Y. Zhu, R.~Zhang, D.~Pathak, T.~Darrell, A.~A. Efros, O.~Wang, and
  E.~Shechtman.
\newblock Toward multimodal image-to-image translation.
\newblock In {\em Advances in Neural Information Processing Systems 30}. 2017.

\end{thebibliography}
}

\newpage
\section*{Appendix}
\section{Implementation details for each task}
The inputs and outputs for each task during the training stage and testing stage are listed in Table \ref{tab:task_io}. 

\begin{itemize}
    \item Image$\to$Image: In this task, the source and target are images drawn from 
    two different domains (e.g. day$\to$night, edges$\to$photos, etc.).
    During training, the references are images drawn from a target distribution, and are 
    paired with the input (i.e. ground truth image). While at testing time reference
    images, which are used for indicating the desired style, can be provided.
    
    \item Text$\to$Image: In this task, the source and reference are paired textual strings and images.
    At testing time, we do not use references to change the images' style or attribute. Because
    a randomly sampled image may change the content that is given by the text. For example. when synthesizing an image from a sentence `This is a little yellow bird', we expect the synthesized image to correctly capture the content of this description. If we were to randomly provide an image with a little white bird, the synthesized birds will have a white appearance. This would cause obvious problems when evaluating the synthesized samples. Therefore, in this task, we only perform synthesis by adding a 
    noise vector, which does not harm the content from the input sentence.
    
    For a convenient implementation, we directly use preprocessed char-CNN-RNN text embeddings
    \cite{reed2016generative} for the
    CUB-200-2011 dataset. This is a method that was also adopted by StackGAN~\cite{stackgan}. Note that, the 
    preprocessed text embeddings are only used for this task. 
    
    \item Image$\to$Text: In this task, the source and reference are paired images and texts. To process text as a reference, we follow the method used in \cite{DBLP:journals/corr/ReedASL16} that allows us to obtain a global sentence text embedding. The input to the reference encoder $Enc_r$ is the average hidden unit activation over the sequence, i.e. $\phi(t)=1 \setminus {L\sum_{L}^{i=1}h_i}$, where $h_i$
    is the hidden activation vector for the $i-th$ frame and $L$
    is the sequence length.  
    
    \item Text$\to$Speech: In this task, we process the speech audio signal as a mel-spectrogram, our model also predicts the mel-spectrogram directly.
    The predicted mel-spectrogram can be synthesized
    directly to speech using either the WaveNet vocoder \cite{WaveNet} or the Griffin-Lim vocoder \cite{griffin-lim}. In our experiments, we use the Griffin-Lim for fast waveform generation. 
    At testing time, the reference speech audio can be from any person (i.e. it does not necessarily need 
    to be sampled from a subject in our training dataset). In our experiments, we test our model by synthesizing 
    audio from reference audio sampled from our dataset, and audio from web (i.e., not from our dataset). It turns out that, the results synthesized by either these references are equivalent in terms of quality and style consistency. We highly recommend readers to 
    hear some of our results on our \href{https://researchdemopage.wixsite.com/cvpr-tts-demo/style-transfer} {demo} page.

    \item Speech$\to$Text: In this task, the speech audio is also processed as a mel-spectrograms. We use the same method as in the Image$\to$Text task to process
    the input reference text as a global sentence embedding. 
    For synthesis, the generator takes a mel-spectrogram sequence combined with the domain embedding obtained from text as input. The output is the predicted text, produced as one character
    at each time step.
    
    \item Text$\to$Text: In this task, the source and reference are textual strings from two languages (e.g., English and French).
    The reference text is processed as a global sentence embedding from the prenet and then fed into a universal attention module to obtain the domain embedding.
    While the source text is output directly from the prenet and is not processed as a global embedding.
    To combine the domain embedding and the source feature, i.e. $(z_r, S)$, 
    the domain embedding $z_r$ is first replicated with $T$ time steps and then concatenated with $S$, where $T$ is the total time step of sequence $S$.
    A detailed illustration of this process can be seen in Figure 5(b) in our paper.
\end{itemize}

\begin{table*}[t]
\caption{Input and output signals in both training and testing stage for each task.}
\begin{center}
    \begin{tabular}{|c|c|c|c|c|c|}
\hline
\multirow{3}{*}{Task} & \multicolumn{2}{c|}{Train}                                                                                                                                    & \multicolumn{3}{c|}{Test}                                                                                                  \\ \cline{2-6} 
                      & \multirow{2}{*}{Input}                                                          & \multirow{2}{*}{Output}                                                     & \multirow{2}{*}{Input}                                                                       & \multicolumn{2}{c|}{Output} \\ \cline{5-6} 
                      &                                                                                 &                                                                             &                                                                                              & $T_{enc}$       & $T_{sam}$       \\ \hline
Image$\to$Image        & \begin{tabular}[c]{@{}c@{}}S: Image\\ R: Image (ground truth)\end{tabular}  & \begin{tabular}[c]{@{}c@{}}(S,R)$\to$$T_{enc}$\\ (S,z)$
\to$$T_{sam}$\end{tabular}  & \begin{tabular}[c]{@{}c@{}}S:Image\\ R:Image (randomly sample)\\ \end{tabular}              & $\checkmark$          & $\checkmark$          \\ \hline
Text$\to$Image         & \begin{tabular}[c]{@{}c@{}}S:Text\\ R:Image (ground truth)\\\end{tabular}     & \begin{tabular}[c]{@{}c@{}} $\setminus$ \end{tabular}  & \begin{tabular}[c]{@{}c@{}}S:Text\\ R:None\\ \end{tabular}                                  &   $\times$       & $\checkmark$        \\ \hline
Image$\to$Text         & \begin{tabular}[c]{@{}c@{}}S:Image\\ R:text (ground truth)\\ \end{tabular}     & \begin{tabular}[c]{@{}c@{}} $\setminus$ \end{tabular} & \begin{tabular}[c]{@{}c@{}}S:Image\\ R:None\\ \end{tabular}                                 & $\times$           & $\checkmark$          \\ \hline
Text$\to$Speech        & \begin{tabular}[c]{@{}c@{}}S:Text\\ R:Speech (mel-spectrogram)\\ \end{tabular} & \begin{tabular}[c]{@{}c@{}} $\setminus$ \end{tabular} & \begin{tabular}[c]{@{}c@{}}S:Text\\ R:Speech(randomly sample)\\ \end{tabular} & $\checkmark$          & $\checkmark$          \\ \hline
Speech$\to$Text        & \begin{tabular}[c]{@{}c@{}}S:Speech\\ R:Text\\ \end{tabular}                   & \begin{tabular}[c]{@{}c@{}} $\setminus$ \end{tabular}   & \begin{tabular}[c]{@{}c@{}}S:Speech\\ R:None\\\end{tabular}                                & $\times$           & $\checkmark$         \\ \hline
Text$\to$Text          & \begin{tabular}[c]{@{}c@{}}S:Text\\ R:Text (ground truth)\\ \end{tabular}      & \begin{tabular}[c]{@{}c@{}} $\setminus$ \end{tabular} & \begin{tabular}[c]{@{}c@{}}S:Text\\ R:None \end{tabular}                                           & $\times$           & $\checkmark$          \\ \hline
\end{tabular}
\end{center}
\label{tab:task_io}
\end{table*}

% \section{Other Application}
% To better validate that our model can learn unified representations from multiple modalities, we tried to make experiments
% to synthesize on task A$\to$C, other than the traditional tasks
% A$\to$B and B$\to$C. To conduct this evaluation, we attempted
% to synthesize from image to speech directly. The model is
% trained on two separate datasets, i.e. MS COCO and EMT-
% 4, which offers image$\to$text and text$\to$speech. In the testing
% stage, the model can directly synthesize from image$\to$speech. It is surprising that the well trained speech synthesize
% model could still generate natural human speech
% from the data which has never seen during training. 

\section{Discussion}

$\bullet$ Why we use the modality subnet for multiple tasks, and
why this makes it easy to add additional tasks.

To use the modality sub-net for multiple tasks
aims to avoid designing different networks for each task.
For example, when we conduct the task of image-to-image
and image-to-text translation, the input modality for both
these tasks are images. In this case, a single image-subnet
could have the potential to produce effective latent code
from all image data. When we want to add more tasks which
start from image modal, this image-subnet can consequently
be used other than grasp a new network.

$\bullet$ Differences with Bicycle-GAN \cite{Bicycle-GAN}

Comparing the task of image-to-image translation with
Zhu et al \cite{Bicycle-GAN}. Our key difference is that UTL allows our model
to encode a large variety of domain information in a latent
space. In addition, the reference encoder learns to produce
representative style code from any given references. In such
a way, we can explicitly control any desired style for synthesizing
by our model, but not just randomly generate diverse
results. Detailed descriptions can be found in the testig
stage of our paper.

\end{document}